# AI-Powered Text Generation for Harmonious Human-Machine Interaction: Current State and Future Directions

Qiuyun Zhang, Bin Guo, Hao Wang, Yunji Liang, Shaoyang Hao, Zhiwen Yu
*School of Computer Science, Northwestern Polytechnical University, Xi'an 710129, P.R.China*
guob@nwpu.edu.cn

*Abstract*—In the last two decades, the landscape of text generation has undergone tremendous changes and is being reshaped by the success of deep learning. New technologies for text generation ranging from template-based methods to neural network-based methods emerged. Meanwhile, the research objectives have also changed from generating smooth and coherent sentences to infusing personalized traits to enrich the diversification of newly generated content. With the rapid development of text generation solutions, one comprehensive survey is urgent to summarize the achievements and track the state of the arts. In this survey paper, we present the general systematical framework, illustrate the widely utilized models and summarize the classic applications of text generation.

## I. INTRODUCTION

Text generation is an important research field in natural language processing (NLP) and has great application prospects which enables computers to learn to express like human with various types of information, such as images, structured data, text, etc., so as to replace human to complete a variety of tasks. The first automatically generated text is dated back to March 17, 2014, when the Los Angeles Times reported the small earthquake occurred near Beverly hills, California by providing detailed information about the time, location and strength of the earthquake. The news was automatically generated by a 'robot reporter', which converted the automatically registered seismic data into text by filling in the blanks in the predefined template text [43]. Since then, the landscape of text generation is rapidly expanding.

At the initial stage, majority studies focused on how to reduce the grammatical errors of the text to make the generated text more accurate, smooth and coherence. In recent years, deep learning achieved great success in many applications ranging from computer vision, speech processing and natural language processing. Most recent advances in text generation field are based on deep learning technology. Not only the most basic Recurrent Neural Networks (RNN) and Sequence to sequence (Seq2seq), but even the Generative Adversarial Networks (GAN) and the Reinforcement learning are widely used in the field of text generation.

With the help of these technologies, the generated text is more coherent, logical and emotionally harmonious. Many dialogue systems have brought great convenience to people's lives such as Microsoft XiaoIce, Contona and Apple siri. They not only help people to accomplish specific tasks, but also communicate with people as a virtual partner. Nowadays, researchers start to consider the research of personalized text generation. Just as we adjust our speaking style according to the characteristics of each other in the daily communication, the text generation process should also dynamically adjust the generation strategy and the final generated content according to the different profiles of the user. Therefore, now the research on personalized text generation is receiving unprecedented attention.

Different from prior survey papers on text generation, in this overview, we introduce the most recent progress from the methodology perspective and summarize the emerging applications of text generation. According to the difference of data modalities, tasks of text generation can be divided into *data-to-text*, *text-to-text*, and *image-to-text*. Among them, data-to-text tasks include weather forecast generation, financial report generation and so on. Text-to-text tasks include news generation, text summarization, text retelling and review generation are widely studied. While, the image-to-text tasks include image captioning, image questioning & answering, etc.

In short, the main contributions of this paper are shown below:

- We summarize the most recent progress in text generation and present the widely used models in this field.

- We provide one comprehensive collection of primary applications including *dialogue systems*, *text summarization*, *review generation* and *image caption & visual question answer*, and the key techniques behind them.

- Finally, we provide a promising research direction of text generation—the *personalized text generation*.

The remaining of this paper is organized as follows. Section 2 introduces the commonly used models in the field of text generation. Section 3 presents the application scenarios of these models in detail. Section 4 highlights application of personalized text generation in various fields. Section 5 summarizes the evaluation and Section 6 concludes this paper with future work.

## II. THE TEXT GENERATION MODELS

In this section, we will introduce the basic frameworks of the widely applied neural networks models for text generation including Recurrent Neural Networks (RNN), Sequence to sequence (Seq2seq), Generative Adversarial Networks (GAN) and Reinforcement learning.

### A. Recurrent Neural Network

RNN is a special neural network structure, which is proposed according to the view that 'people's cognition is based on past experience and memory'. Different from deep neural networks (DNN) and convolutional neural networks (CNN), RNN not only considers the input of the previous moment, but also endows the network with a 'memory' function of the previous content. The RNN structure is shown in Figure 1. RNN can remember the previous information and apply it to the calculation of the current output. Thus, the nodes between the hidden layer are no longer connectionless but connected, and the input of the hidden layer includes not only the output of the input layer but also the output of the hidden layer at the previous moment.

There are also many variations of RNN networks, such as Long Short-Term Memory (LSTM) and Gated Recurrent Unit (GRU). The study in [36] is the pioneering application of RNN for the construction of language models. The experimental results show that the RNN language model outperforms the traditional method.

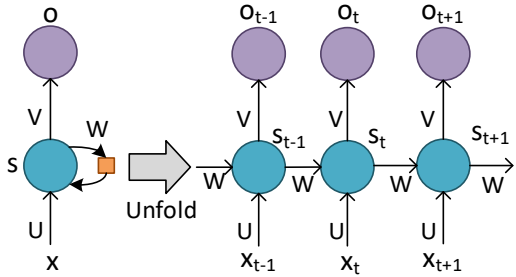

Figure 1. The model structure of RNN

### B. Sequence to sequence structure

Standard seq2seq model used two RNN networks to compose the encoder-decoder structure [7]. The first RNN encoded a sequence of symbols into a fixed length vector representation, and the second RNN decoded the representation into another sequence of symbols. The encoder and decoder were jointly trained to maximize the conditional probability of a target sequence given a source sequence. The Seq2seq structure is shown in Figure 2.

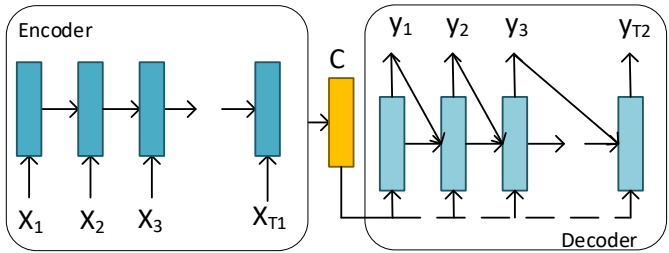

Figure 2. The model structure of Seq2seq

While CNN cannot process sequence data with variable length, and the input and output sequence length of RNN must be the same, using seq2seq model to encode with RNN in encoder stage can receive sequences with indefinite length as input, and in decoder stage can transform the representation vector into sequences without being affected by the input sequence length. Thus, seq2seq is widely used in a variety of tasks including machine translation, text summarization, reading comprehension, and speech recognition, etc.

### C. GAN and Reinforcement learning

GAN[15] proposed by Goodfellow consists of two parts: one generator and one discriminator. The generator is to generate a false sample distribution that is closest to the real samples. The discriminator is used to distinguish generated samples and real samples. The model structure of GAN is shown in Figure 3.

While, the original GAN supports well for the continuous data instead of discrete data such as text. To address this problem, researchers have made some fine-tuning to GAN's structure, which brings hope for the generation of discrete data [2; 22]. Zhang et al. used LSTM as generator and CNN as discriminator to implement the task of text generation [70]. The idea of smooth approximation was used to approximate the output of the generator LSTM to solve the gradient inducibility problem caused by the discrete data.

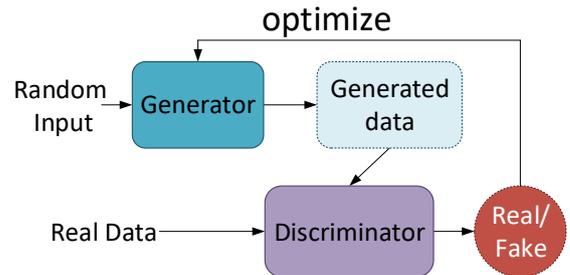

Figure 3. The model structure of GAN

Reinforcement learning is usually a markov decision process in which performing an action in each state will be rewarded (or get negative reward--punishment). The goal of reinforcement learning is to find the optimal policy to maximize rewards. The dialogue generation task is in line with the operating mechanism of reinforcement learning. The dialogue generation process can be seen as a process of maximizing the expected rewards of generated dialogue content.

Through the combination of reinforcement learning and GAN, excellent results have been achieved in the field of text generation. SeqGAN[67] used discriminator in GAN as the source of reward in reinforcement learning. Before the discriminator was updated, the generator continuously optimized itself according to the return score of the current discriminator until the texts generated by the generator were absolutely true. By using the reward mechanism and the policy gradient technologies in reinforcement learning, the problem that the gradient cannot be back propagated when GAN faces discrete data was skillfully avoided. In the interval of training generator with reinforcement learning method, the discriminator was trained with the original method of GAN.

## III. THE TEXT GENERATION APPLICATIONS

In this section, we summarize the classic applications of text generation including dialogue systems, text summarization, review generation and image caption & visual question answer.

### A. Task-oriented Dialogue Systems

Dialogue systems attracted more and more attention in recent years. According to different application fields, dialogue systems can be divided into two categories: task-oriented and non-task-oriented dialogue systems (also known as chatbots).

Task-oriented dialogue systems help users carry out specific tasks, such as restaurant reservations, travel itineraries, etc. Apple Siri and Microsoft Cortana are the representatives of the task-oriented dialogue system. Recently, deep learning algorithm has been applied to the construction of task-oriented dialogue systems. Deep learning can automatically learn high-dimensional distributed feature representation and reduce the burden of manual design. Using a large amount of dialogue data to build a pure data-driven end-to-end dialogue system to directly map user input to system output is a very popular research direction now.

Wen et al. [62] constructed a task-oriented dialogue system by using a modular neural generation model. Neural network is used to realize the process of all modules, and specific tasks of restaurant reservation was achieved. Bordes et al. used the neural generation model to treat the dialogue process as a mapping between the user input content and the model reply content, and used the encoder-decoder structure to train the mapping relationship [4]. In order to solve the problem of dependence on external knowledge bases in task-oriented dialogue systems, Eric et al. proposed the end-to-end key-value retrieval network in [11], which was equipped with an attention-based key-value retrieval mechanism over entries of a knowledge base, and could extract relevant information from the knowledge base. In addition, the memory network, a variant of RNN, was proposed to store the current user's dialogue context and similar user's conversation history with external memory module [31]. By matching user input and context, appropriate replies could be selected from the alternative reply set.

### B. Non-task-oriented Dialogue Systems

Known as chatbots, non-task-oriented dialogue systems aim to communicate with humans naturally in the open context. Microsoft XiaoIce is a typical chatbots. There are two main design methods for the non-task-oriented dialogue systems: retrieval-based method and generative method.

#### 1) Rtrieval-based methods

The retrieval-based method directly selects the corresponding reply from the alternative replies of the given corpus according to the matching principle. In [29], dual encoder model was proposed by Lowe et al. for semantic representation of context and reply content. Context and reply were respectively encoded into semantic vectors by dual RNN model, and then semantic similarity was calculated by matrix transformation. It was found that matching only from the perspective of words could not achieve good results, so Zhou et al. proposed matching through multiple levels (word level and utterance level), and its multi-dimensional thinking provided direction for the following papers [71]. In [72], Zhou et al. used the encoder part of the transformer model obtain the multi-granularity text representation of each context and reply, and then two matching matrices were calculated for the representation under each granularity of utterance-response pair, and the dependency information between the words in utterance and the words in response was also added to the calculation of the alignment matrix as the expression of the words, so as to model a deeper semantic relationship.

#### 2) Generative methods

Recently, the data-driven model has been widely studied in the dialogue system. The pure data-driven model directly trains from a large amount of dialogue data without relying on external knowledge. Ritter et al. [49] took the reply generation problem as a translation problem, in which the process of generating replies was regarded to translate the query into corresponding replies. Based on the statistical machine translation model, a generating probability model was proposed to model the dialogue system. The disadvantages of that model are obvious. The most important one is that only one user query is translated into the reply in the translation process without considering the context information in the dialogue, which is obviously unable to work properly in multiple rounds dialogues.

With the development of deep learning, neural generation model began to receive attention. Sordoni et al. and Vinyals et al. [53; 59] began to apply RNN to construct the dialogue model, and applied the neural network method to the end-to-end dialogue model for the first time. Based on the Seq2Seq model, the past dialogue history was mapped to the reply. Many existing researches have realized the importance of context. The simplest method is to use RNN to directly encode the dialogue sentences as a whole sequence, and obtain the semantic representation vector of the context, which is treated as additional input in the decoding stage. This method is used by Yan et al. [64] to utilize the context information. Direct concatenation of all sentences may lose the relative relationship between sentences, so researchers have proposed more complex methods to extract context information. Using the multi-layer model to extract context information, the first level is a model of the sentence level to encode the semantic information of a single sentence, and the second is a cross-sentence level model, using the first layer's output as input, to integrate all the contextual information. Tian et al. [57] carried out experiments on three different cross-sentence methods and came to the conclusion that the performance of the multi-layer context information extraction model outperformed the single-layer model.

As stated in [63], in daily human communication, people often associate a dialogue content with related topics in their mind. Based on this assumption, Xing et al. organized content and selected words according to the topics for generating responses. The Latent Dirichlet Allocation (LDA) topic model was used to obtain the topical information in the dialogue sentences, which was taken into consideration as additional input in the decoding process. The experimental results showed that the introduction of topics into the dialogue model is constructive with improved performance. Choudhary et al.

Recently, GAN and reinforcement Learning have been applied to dialogue systems. Li et al. through combining GAN and reinforcement learning, jointly trained two models [26]. The generation model aimed to generate reply sequences, and the discriminator was used to distinguish between human-generated and machine-generated dialogue. Li et al. [25] simulated dialogues between two virtual agents, using policy gradient to reward sequences showing three useful dialogue attributes: informativity, coherence, and ease of answering (related to forward-looking function).

### C. Text Summarization

Text summarization is another important research direction in text generation which provides concise description for users by compressing and refining the original text. Text summarization can be regarded as a process of information synthesize, in which one or more input documents are integrated into a short abstract. Banko et al. viewed summarization as a problem analogous to statistical machine translation and generated headlines using statistical models for selecting and ordering the summary words [3]. There are two methods to realize text summarization: retrieval-based method and generative method which will be described in detail in the rest of this chapter.

#### 1) Retrieval-based methods

Retrieval-based method is a simple method by selecting a subset of sentences in the original document. This process can be thought of selecting the most central sentences in the document, which contain the necessary and sufficient information related to the subject of the main theme.

Nenkova et al. [40] used the word frequency as a feature of the summarization. Three attributes related to word frequency were studied: word frequency, compound function estimating sentence importance from word frequency, and word frequency weight adjusted based on context. Erkan et al. [12] proposed a model based on the centrality (prestige) of eigenvectors, which is known as LexPageRank. This model constructed the sentence connectivity matrix based on cosine similarity.

Svore et al. [55] proposed a new automatic summarization method based on neural network, whose name was NetSum. The model retrieved a set of characteristics from each sentence to help determine the importance in the document. In the [6], Cheng et al. used neural network to extract abstract, and word and sentence contents were extracted respectively. What is special about this work is the use of the attention mechanism. They directly used the scores in attention to select sentences in a document, and was actually similar to pointer networks. Cao et al. [5] used the attention mechanism to weight the sentences. The weighted basis was the correlation of document sentences to query (based on attention), and thus extracted the summary by ranking the sentences.

The disadvantages of retrieval-based method include the similarity of selected sentences and the lack of logic among the selected sentences.

*2) Generative methods*

Different from retrieval-based method, the generative method is able to generate sentences that are not in the original text, which requires the generative model to have stronger ability of understanding and representation. It is difficult for traditional methods to achieve these abilities.

Paulus et al. [45] introduced the application of reinforcement learning method based on Seq2Seq architecture in abstract generation. Pasunuru et al. also used reinforcement learning to generate the summarization of the article [44].

The theme of [50] from Facebook was attention-based NN to generate sentence summarization. Alexander M. Rush et al. proposed a sentence digest model under the framework of encoder-decoder. Later, this method was used in many works to construct training data. Nallapati et al. [39] not only included work on sentence compression, but also presented a new data set about document into a multi-sentence. This paper added a lot of features on it, such as POS tag, TF, IDF, NER tag, etc. The feature-rich encoder proposed in this paper was also of great significance for other work.

*D. Review Generation*

Review generation belongs to data-to-text natural language generation [14]. Within the field of recommender systems, a promising application is to estimate (or generate) personalized reviews that a user would write about a product, to discover their nuanced opinions about each of its individual aspects [41]. In order to recommend products to users, we need to ultimately predict how users will react to new products. However, traditional methods often discard comment text, which makes the underlying dimensions of users and products difficult to explain [35].

Huang et al. [16] determined the meaning of words through the context of local and global documents of words and explained homonyms and polysemy by learning multiple embedding of each word. Tang et al. [56] generated natural language in a specific context or context, which introduced two text generation models, C2S and gC2S. The C2S model produced semantically and syntactically coherent sentences, and gC2C did better when the sequence became very long. After identifying the product domain, name, and user rating, the model could generate a review of the corresponding rating like "I love Disney movies but this one was not at all what I expected. The story line was so predictable, it was dumb." Jaech et al. [17] made full use of RNN and concatenated the context with the word embedding at the input layer of RNN. Experiments on language modeling and classification tasks using three different corpora demonstrated the advantages of this method.

Almahairi et al. [1] developed two new models (BoWLF and LMLF) to normalize the rating predictions for the Amazon review data set using text reviews. Lei Zheng et al. [23] proposed a new method for modeling ratings, reviews and their temporal dynamics in conjunction with RNN was proposed. A recurrent network was used to capture the temporal evolution of user and movie states, which were directly used to predict ratings. The user's movie rating history was used as the input of the updated status.

The problem with review generation is how to use fine-grained attributes as input to generate more diverse and user-specific comments. Generation of long comments is also a challenge.

*E. Image Captioning & Visual Question Answering*

With the development of social network, the task that generates captions for images received a lot of attention.

*1) Image Captioning*

Image caption is a basic multimodal problem in the field of artificial intelligence, which connects computer vision with natural language generation. It can be divided into two steps, feature extraction and natural language generation. CNN is usually used as the feature extraction sub-model. It can extract significant features, usually represented by the context vector of fixed length. This is followed by a RNN model to generate the corresponding sentence. The whole structure is similar to encoder-decoder structure.

Jaech et al. proposed a deep Boltzmann machine in [54] to learn how to generate such multimodal data and shows that the model can be used to create a fused representation by combining features across modes. Kiros et al. [19] introduced the neural language model of multimodal constraint and used CNN to learn the word representation and image features together. Vinyals et al. [60] proposed a generation model based on deep RNN architecture. Given the training image, the model could be trained to maximize the probability of the target sentence.

Socher et al. introduced a model that recognized objects in images even when there was no training data available in the object class [52]. And in the completely unsupervised model, the accuracy was up to 90%. Mao et al. [34] proposed a multimodal RNN (m-RNN) model to generate new sentence descriptions explaining the content of images. The model was composed of two subnetworks: sentence depth RNN and image depth CNN.

Kulkarni et al. [21] introduced an automatic natural language description generation system based on image, which used a lot of statistical information of text data and computer vision recognition algorithm. The system was very effective in generating image-related sentences. Mitchel et al. [37] used a new method to generate language, in which syntactic models were linked to computer vision detection to generate well-formed descriptions of images by filtering out unlikely attributes and putting objects into ordered syntactic structures. Frome et al. [13] proposed a new deep visual semantic embedding model which used annotated image data and semantic information extracted from unannotated text to identify visual objects.

*2) Visual Question Answering*

Visual question answering (VQA) aims to answer questions about image. The inputs are one image and one question associated with the image, and the output is one answer to the question. The deep learning model of VQA usually uses CNN to acquire image's information and RNN to encode the question.

Ma et al. applied CNN to VQA tasks [32] and provided an end-to-end convolutional framework for learning not only images and problem representations, but also the modal interactions between them to generate answers. Malinowsk et al. [33] also used CNN to encode image and feed the question together with the image representation into the LSTM network. The system was trained to give correct answers to questions about images. Ren et al. [48] used neural networks and visual semantic embedding and not included intermediate stages, such as object construction and image segmentation.

There are other methods besides CNN to implement VQA task. Noh et al. [42] used an independent parametric predictive network with a GRU with the question as input and a fully connected layer generating as output. By combining hashing techniques, they reduced the complexity of constructing a parameter prediction network with a large number of parameters. Yang et al. [66] proposed a learning method based on hierarchical attention network, which could help models to answer natural language questions from images. Zhu et al. [73] evaluated several basic patterns of personnel performance and QA tasks and proposed a new LSTM model with spatial attention that can handle 7W quality assurance tasks.

## IV. PERSONALIZED TEXT GENERATION

With the development of deep learning techniques, we hope computers to automatically write high-quality natural language text, but much of the previous research has focused on the generated text content, not the user's personality. In our daily conversation, we will not only consider the fact content to produce the corresponding dialogue content, but also consider other's personalized profiles, to adjust our dialogue style and strategy. The purpose of personalized text generation is to let the computer imitate the behavior of human beings and take the personalized characteristics of users into consideration when generating text content, so as to dynamically adjust the generated text content and generate more high-quality text. Personalized text generation are embodied in many applications, which will be briefly introduced below.

### A. Personalized Dialogue Systems

Dialogue system is one of the text generation applications that best reflects the user's personalized profiles. In the process of chatting with different users, if chatbots want to bring pleasant interactive experience to users, it needs to adjust its dialogue strategies and reply content according to different characteristics of users.

In [24], the personalized characteristics of the user were modeled for the first time by Li et al., and the persona-based model was proposed. Different user was embedded into the hidden vector space by the similar word embedding method. The user embedding vector was used to adjust the dialogue style and content of the dialogue agent. Kottur et al. [20] extended the previous model, and also carried out vector embedding of user features. Combined with the Hierarchical Recurrent Encoder-Decoder structure, the model could better capture context-related information and considered user personalized features to generate more high-quality dialogue content.

Considering the lack of dialogue data with user personalized characteristics, Luan et al. [30] applied the multi-task learning mechanism to the personalized reply generation. A small amount of personalized dialogue data was used to train the reply generation model firstly, and then an auto-encoder model was trained with non-conversational data and the parameters of the two models were shared by the multi-task learning mechanism to obtain the generation model of personalized reply. Mo et al. and Yang et al. made use of the idea of transfer learning [38; 65]. First, they trained a large number of general dialogue data to generate a general reply model, and then used a small amount of personalized dialogue data to fine-tune the model with transfer learning, so that users' personalized information could be considered when generating reply.

Considering the different influence of user characteristics on the reply content, Qian et al. [46] applied the supervision mechanism to judge when to express the appropriate user profiles in the reply generation process. Liu et al. [28] built a two-branch neural network to automatically learn user profiles from user dialogues, and then the deep neural network was used to further learn fusion representation the user queries, replies and user profiles, so as to realize the dialogue process from the user's perspective. Zhang et al.

All the work above was carried out based on chatbots, and the task-oriented dialogue system is also an important research direction in the dialogue system, but there are few studies that consider the user's personalized information in it. Joshi et al. [18] published the dataset of task-oriented dialogue system, in which each conversation contained the user's personalized information, providing data support for subsequent research. Luo et al. [31] made use of a variant of RNN--memory network to realize the task-oriented personalized dialogue system. The Profile Model was used to encode the user's personalized information, and the Preference Model was used to solve the ambiguity problem of the same query when facing different users. At the same time, the similar user's dialogue history was stored. When the reply content was extracted, the personalized reply content for different users was generated by combining the similar user dialogue history and the user personalized feature information.

### B. Personalized Review Generation

The findings of Tintarev et al. [58] indicated that users mentioned different movie features when describing their favorite movies and short, personalized arguments for users were more persuasive. Therefore, personalized user review generation contributes to better recommend products.

Radford et al. [47] demonstrated the direct influence of emotional units on the process of model generation. Lipton et al. [27] built a system of giving user/item combinations to generate the comments that users would write when reviewing the product. They designed a character-level RNN to generate personalized product reviews. The model learned the styles and opinions of nearly a thousand different authors, using a large number of comments from BeerAdvocate.com. The model in [9] was able to generate sentences which was close to a real user's written comments and could identify spelling errors and domain-specific words.

Besides RNN, the decoder structure can be LSTM and GRU. Zang et al. [68] introduced a deep neural network model to generate Chinese comments from emotional scores representing user opinions. In this paper, a hierarchical (LSTM) decoder with attention consistency was proposed. Dong et al. [10] proposed an attention-enhanced attribute-to-sequence model to generate product reviews for given attribute information, such as users, products, and ratings. Attribute encoder learned to represent input attributes as vectors. The sequence decoder then generated comments by adjusting the output of these vectors. They also introduced an attention mechanism to syndicate comments and align words with input attributes.

Sharma et al. [51] used the model similar to [10] and added loss terms to generate more compliant comments. Ni et al. [41] designed a review generation model that could make use of user and project information as well as auxiliary text input and aspect perception knowledge. In the encoding stage of the model, there were three encoders (sequence encoder, attribute encoder and aspect encoder) for information integration. The decoder's processing of the encoded information biased the GRU model toward generating phrases and statements closest to the input.

In addition, GAN can also be used to generate personal reviews. Wang et al. [61] proposed a new punishment-based goal that took a more rational approach to minimizing overall punishment rather than maximizing rewards. Experiments and theories have shown that based on the punishment, it could force each generator to produce multiple texts of specific emotional tags, rather than producing repeated but "safe" and "good" examples. In addition, multi-class discriminator target allowed the generator to focus more on generating its own specific emotional tag examples. The model could generate a variety of different emotional tags of high-quality text perceptual language.

## V. EVALUATION METRICS

With the continuous development of text generation technology, the corresponding evaluation method has gradually become an active research direction. Researchers need to use the established evaluation method to judge the quality of the proposed models. Good evaluation metric is a key factor to promote the research progress. To date, there are two main methods for text generation evaluation: objective evaluation metric and artificial evaluation. Objective evaluation metric is mainly divided into two aspects, the first is the word overlapping evaluation matrix, such as BLEU and ROUGE, the second is based on word vector evaluation matrix, such as Greedy Matching and Embedding business.

### A. Word overlap evaluation metrics

#### 1) BLEU (bilingual evaluation understudy)

BLEU is a method to compare the n-gram of model output and reference output and calculate the number of matched fragments. To calculate this metric, you need to use machine-translated text (called candidate docs) and some text translated by professional translators (called reference docs). In essence, BLEU is used to measure the degree of similarity between machine translation text and reference text. Its value ranges from 0 to 1, and the closer the value is to 1, the better the machine translation results will be.

BLEU adopts the n-gram matching rule, through which it calculates a proportion of N groups of words similar between the comparison translation and the reference translation. BLEU algorithm can give relatively valuable evaluation scores quickly.

#### 2) ROUGE (Recall-Oriented Understudy for Gisting Evaluation)

ROUGE evaluates the abstract based on the co-occurrence information of n-gram, which is an evaluation method oriented to the recall rate of n-gram words. Its basic idea is to generate an abstract set of standard abstracts by a number of experts, respectively, and then to compare that automatically generated abstract of the system with the artificially generated standard abstract. The quality of abstract is evaluated by counting the number of overlapping basic units (n-gram, word sequence and word pair) between the two kind of abstract. The stability and robustness of the system are improved by comparing the multi-expert manual abstracts. This method has become one of the general notes of abstract evaluation technique.

### B. Word vectors evaluation metrics

In addition to the word overlap, another way to evaluate the response effect is to judge the relevance of the response by knowing the meaning of each word, and the word vector is the basis of this evaluation method. In accordance with semantic distributions, a vector is assigned to each word by a method such as word2vec, which is used to represent that word, which is represented approximately by calculating the frequency that the word appears in the corpus. All the word vector metrics can be approximated as sentence vectors at the sentence level through vector connection. In this way, the sentence vectors of candidate and target reply sentences can be obtained respectively, and the similarity between them can be obtained by comparing them with cosine distance.

#### 1) Greedy matching

The greedy matching method is a matrix matching method based on word level. In the two sentences given $r$ and $\hat{r}$, every word $w \in r$ will converse into word vector $e_w$ after a conversion. At the same time, cosine similarity matching is carried out to the maximum extent with the word vector $e_{\hat{w}}$ of each word sequence $\hat{w} \in \hat{r}$ in $\hat{r}$, and the final result is the mean value after all words are matched. Greedy matching was first proposed in the intelligent navigation system, and subsequent studies have found that the optimal solution of this method tends to be the result with a large semantic similarity between the center word and the reference answer.

#### 2) Embedding average

The embedding average is the way to calculate a sentence eigenvector by the word vector in the sentence. The vector of a sentence is calculated by averaging the vectors of each word in the sentence. It's a method that has been used in many NLP domains beyond the dialogue system, like calculating the similarity of the text. When comparing two sentences, embedding average can be calculated respectively and then put the cosine similarity of both as indicators to evaluate their similarity.

## VI. FUTURE DIRECTIONS AND CONCLUSION

Within the last two decades, although great achievements of deep learning spur the development of text generation, it is still at the preliminary stage with a large number of open issues. In this section, we will highlight some pending questions to underpin the future of research work.

### A. Dataset Deficiency

Different from computer vision and machine translation, there is a lack of high-quality data in the field of text generation and it is difficult to manually label data. How to use a small amount of data to complete the efficient training of the model is the primary research direction in the future.

### B. Ultra-long Dialogue Context

Human-computer dialogue is a hot area of text generation research. Although current chatbots can preliminarily understand the context, it is still difficult to grasp the ultra-long text. How to effectively capture the semantic information in the text and ensure the consistency of language and logic in the whole dialogue process is a hot topic in the future.

### C. Co-textual Information

Most of the existing studies only focus on the text content, but ignore their co-textual information. However, in reality, natural language is usually generated in a specific environment, such as time, place, emotion or emotion. Therefore, only by considering these co-textual information can the syntactically correct, semantically reasonable and reasonable text content in a specific context be generated.

### D. Evaluation Metrics

Text generation field is lack of unified evaluation metrics system, the best evaluation method is conducted by artificial judgement. High-quality evaluation metric is crucial to the research in the field of artificial intelligence. Only through reasonable and unified evaluation metric can researchers know whether their research work is reasonable or not. This is a major research gap in the future.

### E. Personalized text generation

Research on personalized text generation is attracting more and more attention. Most of the existing research is based on the encoding of user personalized profiles. How to effectively obtain the relationship between personalized profiles and text content is the focus of future research. Another problem is the impact of the lack of personalized data on model training. How to use a small amount of personalized data to achieve personalized text generation is the focus of researchers.

This paper gives a comprehensive introduction to the basic concepts, commonly used models, and popular applications in the text generation. At the same time, some unsolved problems are put forward. Since there are many researchers in each of these work, the relevant research results are also endless, so the text is inevitably missing. We hope that this paper will provide some help to relevant researchers in this field.

## ACKNOWLEDGMENT

This work was partially supported by the National Key R&D Program of China(2017YFB1001800), the National Natural Science Foundation of China (No. 61772428, 61725205).